\documentclass[a4paper,fleqn]{cas-dc}

\usepackage[numbers]{natbib}

\def\tsc#1{\csdef{#1}{\textsc{\lowercase{#1}}\xspace}}
\tsc{WGM}
\tsc{QE}
\tsc{EP}
\tsc{PMS}
\tsc{BEC}
\tsc{DE}

\begin{document}
\let\WriteBookmarks\relax
\def\floatpagepagefraction{1}
\def\textpagefraction{.001}
\shorttitle{QC-SMOTE}
\shortauthors{P. Upman and S. N. Gowda}

\title [mode = title]{QC-SMOTE: Quality-Controlled SMOTE for Imbalanced Classification}                      




\author{Parth Upman}
\ead{psxpu1@nottingham.ac.uk}

\credit{Conceptualization of this study, Methodology, Software}

\author{Shreyank N Gowda}
\cormark[1]
\ead{shreyank.narayanagowda@nottingham.ac.uk}

\affiliation{organization={School of Computer Science, University of Nottingham},
                addressline={Jubilee Campus, Wollaton Road}, 
                city={Nottingham},
                postcode={NG8 1BB}, 
                country={United Kingdom}}

\cortext[cor1]{Corresponding author}

\begin{abstract}
Class imbalance poses a significant challenge in classification, where existing methods such as SMOTE often generate low-quality synthetic samples in regions with noise or class overlap. We propose QC-SMOTE, a quality-controlled oversampling framework that estimates minority sample reliability using a composite neighbourhood trustworthiness score combining local density, safe-level, and isolation from the majority class. Synthetic candidates are generated using an IPQ-guided best-of-K strategy that evaluates midpoint purity and, when required, majority clearance, with allocation guided by sample reliability and boundary informativeness. Generation behaviour adapts across overlap--imbalance regimes, adjusting interpolation range and selection criteria to match local data geometry. Low-quality synthetic samples are replaced with original minority duplicates when neighbourhood purity falls below an adaptive threshold, providing graceful degradation by reverting to duplication in severely noisy regions. Experiments on 30 imbalanced datasets using repeated stratified cross-validation show that QC-SMOTE achieves the strongest average AUC-ROC and Macro F1 among the compared oversampling methods, with particularly clear gains under moderate and severe imbalance. These results demonstrate the importance of quality-aware, geometry-adaptive synthetic sampling for robust imbalanced classification.
\end{abstract}


\maketitle

\section{Introduction}

Many real-world classification problems are defined not by the abundance of data, but by the scarcity of the examples that matter most. In fraud detection, medical diagnosis, fault prediction, and anomaly detection, the rare class often corresponds to the event of greatest practical importance \cite{dal2017credit,khalilia2011predicting,sowjanya2023effective,zhou2022distribution}. However, standard classifiers trained on imbalanced data tend to favour the majority class, since doing so can yield high overall accuracy while still failing to recognise minority patterns. This creates a mismatch between conventional empirical performance and the actual objective of many applications such as the reliable recognition of rare but consequential events.

\begin{figure*}[t]
    \centering
    \includegraphics[width=\textwidth]{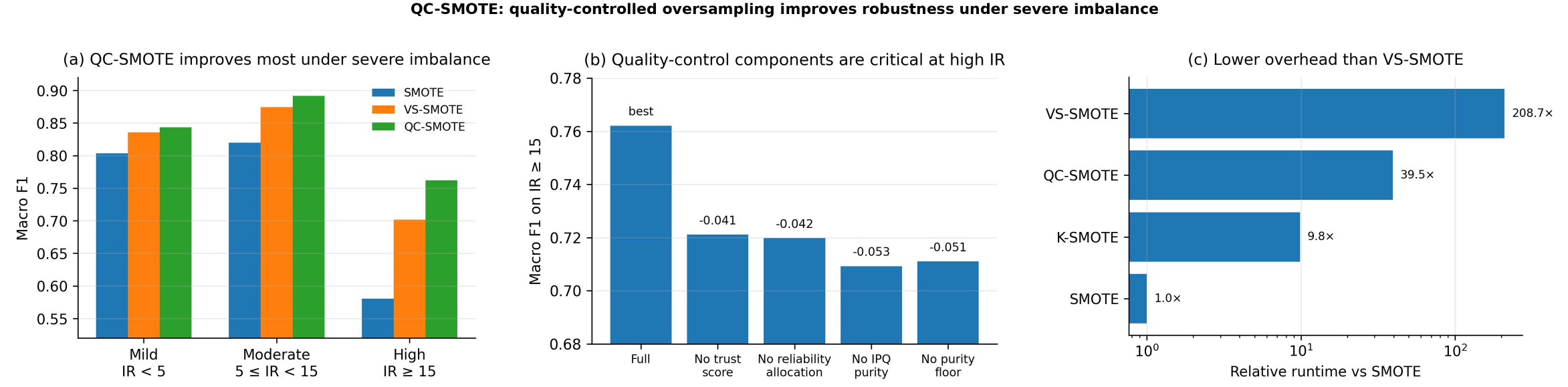}
    \caption{\textbf{Teaser summary of QC-SMOTE.} The proposed method improves class-balanced performance across mild, moderate, and high imbalance regimes, with the largest gains under severe imbalance. Component ablations show that trustworthiness estimation, reliability-weighted allocation, IPQ-guided candidate selection, and purity filtering are particularly important in high-imbalance settings. QC-SMOTE also provides a favourable runtime--performance trade-off, remaining substantially more efficient than VS-SMOTE while improving AUC-ROC and Macro F1.}
    \label{fig:teaser}
\end{figure*}

Approaches to imbalanced learning are commonly grouped into data-level, algorithm-level, and cost-sensitive methods \cite{he2009learning,fernandez2018learning,krawczyk2016learning,udu2025emerging,fernandez2018smote}. Algorithm-level methods modify the learner, for example through ensemble design or class reweighting, while cost-sensitive methods penalise minority-class errors more heavily. These strategies can be effective, but they are often tied to particular model families or require careful tuning of loss weights. Data-level methods instead act directly on the training distribution, making them attractive because they are simple, classifier-agnostic, and easy to integrate into existing machine-learning pipelines.

Among data-level methods, the Synthetic Minority Over-sampling Technique (SMOTE) \cite{chawla2002smote} remains one of the most influential and widely used approaches. SMOTE generates new minority samples by interpolating between existing minority instances, thereby expanding minority support without simply duplicating observations. This simple idea has motivated a large family of variants, including boundary-aware, density-aware, clustering-based, and noise-filtered extensions \cite{han2005borderline,camacho2022geometric,sun2024smote,arafa2022rn,bunkhumpornpat2009safe,qiu2025vs}. Despite this progress, the central operation of many SMOTE-based methods still depends on a strong geometric assumption that interpolation between nearby minority points is usually safe.

In practice, this assumption is often violated. Minority samples may lie near class boundaries, inside overlapping regions, or as isolated outliers. Interpolating from such points can create synthetic samples that are ambiguous, noisy, or closer to the majority distribution than to the true minority structure. Boundary-focused methods such as Borderline-SMOTE \cite{han2005borderline} and ADASYN \cite{he2008adasyn} attempt to place more emphasis on difficult regions, while density- and clustering-based approaches \cite{douzas2018improving} seek more informative regions for generation. However, these approaches still face a difficult trade-off. Focusing too strongly on boundary or sparse regions can amplify noise, while concentrating only on safe dense regions may reduce diversity and leave parts of the minority distribution underrepresented.

This suggests that effective oversampling requires more than deciding \emph{where} minority samples are located. It also requires deciding \emph{which} minority samples are trustworthy, \emph{where} interpolation is geometrically reliable, and \emph{how} generation should change as the data move from mild imbalance to severe imbalance and overlap. Recent methods have begun to recognise this issue by evaluating interpolation spaces, incorporating density information, or using majority-neighbourhood cues \cite{qiu2025vs,khorshidi2025somm,hemmatian2025crn,widiyaningtyas2025nrmodified}. Nevertheless, most existing methods remain organised around a single dominant principle, such as boundary emphasis, clustering, density, or post-hoc cleaning. As a result, they do not fully couple sample-level reliability, candidate-level quality assessment, and regime-adaptive generation within a unified oversampling process.

To address this gap, we propose \textbf{QC-SMOTE}, a quality-controlled oversampling framework for imbalanced classification. The key idea is that synthetic samples should not be accepted merely because they lie between two minority points. Instead, QC-SMOTE first estimates the reliability of minority samples using a composite neighbourhood trustworthiness score that combines local density, safe-level characteristics, and isolation from the majority class. This score guides seed selection and synthetic-sample allocation, giving more generation capacity to minority samples located in reliable regions of the feature space.

QC-SMOTE then introduces an \emph{interpolation purity and quality} (IPQ)-guided best-of-$K$ generation strategy. Rather than producing a single synthetic sample from each seed--neighbour pair, the method generates multiple candidates and evaluates their local purity and, when required, their clearance from the majority class. This candidate-level quality control allows QC-SMOTE to reject poor interpolation outcomes before they are added to the training set. In addition, the method adapts its generation behaviour across overlap--imbalance regimes, adjusting interpolation range, candidate selection, and fallback behaviour according to the geometry of the data. When no reliable synthetic candidate is available, QC-SMOTE degrades gracefully by reinforcing reliable minority samples through duplication rather than injecting potentially harmful synthetic points.

Figure~\ref{fig:teaser} summarises the empirical motivation and benefits of the proposed framework. QC-SMOTE improves both AUC-ROC and Macro F1 across mild, moderate, and high imbalance regimes, with the largest improvements appearing in the high-imbalance setting where minority examples are sparse and interpolation is most fragile. The ablation results further show that trustworthiness estimation, reliability-weighted allocation, IPQ midpoint purity, and purity filtering are especially important under severe imbalance. At the same time, QC-SMOTE maintains a favourable runtime profile, remaining substantially more efficient than VS-SMOTE while improving the main evaluation metrics.

We evaluate QC-SMOTE on 30 real-world imbalanced datasets using repeated stratified cross-validation. The proposed method is compared against a broad set of SMOTE-based baselines, including recent oversampling methods. Across the benchmark, QC-SMOTE achieves the strongest average AUC-ROC and Macro F1, with particularly clear gains in moderate- and high-imbalance regimes. These results support the central hypothesis of this work: oversampling is most effective when synthetic generation is guided jointly by sample reliability, candidate quality, and the local imbalance--overlap regime.

These findings also highlight a broader point about synthetic oversampling that performance gains do not come only from increasing the number of minority samples, but from controlling the reliability of the regions in which those samples are generated. In this sense, QC-SMOTE shifts the emphasis from quantity-driven oversampling to quality-aware data construction. This distinction is especially important in high-imbalance settings, where a small number of unreliable synthetic samples can disproportionately affect the learned decision boundary. By combining reliability estimation, candidate filtering, and regime adaptation, QC-SMOTE aims to make synthetic oversampling more dependable across heterogeneous data conditions.

The main contributions of this work are summarised as follows:
\begin{itemize}
    \item We introduce \textbf{QC-SMOTE}, a quality-controlled oversampling framework that explicitly models minority-sample reliability using a composite neighbourhood trustworthiness score.

    \item We propose an \textbf{IPQ-guided best-of-$K$ candidate selection mechanism} that evaluates the quality of candidate interpolation regions before accepting synthetic samples into the training set.

    \item We develop a \textbf{regime-adaptive generation strategy} with a graceful degradation mechanism, allowing the method to adjust its behaviour in noisy, sparse, or highly overlapping regions.

    \item Together, these components provide a unified and practical framework for robust synthetic oversampling in imbalanced classification.
\end{itemize}

\section{Related Work}
\label{sec:related_work}

\subsection{Imbalanced Learning and Data-Level Resampling}

Class imbalance has long been recognised as a central challenge in supervised learning, particularly in applications where the minority class corresponds to rare but important events such as fraud, disease, system failure, or anomalous behaviour \cite{he2009learning,fernandez2018learning,krawczyk2016learning}. In such settings, standard classifiers often favour the majority class, producing high overall accuracy while failing to identify the minority cases that matter most. Recent surveys emphasise that imbalance rarely appears in isolation: it is frequently coupled with class overlap, noisy labels, small disjuncts, heterogeneous density, and domain-specific constraints \cite{chen2024survey,carvalho2025resampling,nikpour2026review}. These factors make imbalance a geometric and distributional problem, rather than a simple issue of unequal class counts.

Methods for imbalanced learning are commonly grouped into data-level, algorithm-level, cost-sensitive, and hybrid approaches. Algorithm-level methods modify the learner or training objective, while cost-sensitive methods assign larger penalties to minority-class errors. Data-level methods instead alter the training distribution through oversampling, undersampling, or a combination of both \cite{he2009learning,fernandez2018learning}. Among these, oversampling is especially attractive because it is classifier-agnostic and can be integrated into standard learning pipelines without changing the downstream model. However, its effectiveness depends critically on whether the added samples improve minority representation without increasing ambiguity near the class boundary.

\subsection{SMOTE and Reliability-Aware Seed Selection}

The Synthetic Minority Over-sampling Technique (SMOTE) \cite{chawla2002smote} is the foundation of many data-level imbalance methods. SMOTE generates synthetic minority samples by interpolating between a minority instance and one of its minority neighbours. This simple mechanism expands minority support without discarding majority data or merely duplicating existing minority samples. However, SMOTE also makes a strong assumption: that nearby minority samples define a reliable region for interpolation. This assumption can fail in the presence of noise, outliers, class overlap, or fragmented minority subclusters.

Many SMOTE variants therefore focus on improving the selection of seed samples. Borderline-SMOTE \cite{han2005borderline} and ADASYN \cite{he2008adasyn} prioritise difficult or boundary-adjacent minority samples, aiming to improve discrimination near the decision boundary. Safe-Level-SMOTE \cite{bunkhumpornpat2009safe}, in contrast, favours minority samples surrounded by other minority neighbours, thereby reducing the risk of generating ambiguous points. Other importance-weighted and density-aware variants similarly attempt to estimate which minority samples are informative, reliable, or underrepresented. These methods established an important principle: oversampling should not treat all minority samples equally.

Nevertheless, seed-level reliability alone is not sufficient. A minority sample may appear useful, but interpolation from that sample can still cross an overlapping region or produce a candidate that lies too close to the majority class. This limitation motivates methods that move beyond asking only which original samples should be selected, toward asking whether the generated candidate itself is reliable.

\subsection{Geometry-Aware and Space-Aware Oversampling}

A second line of work modifies the geometry of synthetic sample generation. KMeans-SMOTE \cite{douzas2018improving} uses clustering to allocate generation across minority regions, helping avoid unsafe clusters while maintaining coverage of minority modes. Geometric SMOTE \cite{douzas2019geometric} generalises interpolation beyond a simple line segment by defining a geometric sampling region around minority samples. Other distance-aware, density-aware, and feature-aware methods adapt the interpolation rule to better reflect local data structure \cite{camacho2022geometric,matharaarachchi2024extsmote,wang2025resampling}.

Recent work has increasingly emphasised the quality of the interpolation space itself. VS-SMOTE \cite{qiu2025vs}, for example, evaluates multiple interpolation subspaces using representative middle samples, explicitly recognising that the region between two minority points may be more important than the endpoints alone. This is a significant conceptual shift: the central object of oversampling is no longer only the original minority sample, but the candidate region in which synthetic data are placed. Theoretical analyses of SMOTE-induced distributions have also shown that synthetic samples can distort the minority distribution even when interpolation is local \cite{elreedy2024distribution}. These findings reinforce the need for explicit quality control during generation.

QC-SMOTE follows this space-aware perspective, but differs by coupling candidate-level quality evaluation with reliability-weighted allocation and regime-adaptive behaviour. Rather than selecting a seed and accepting a single interpolation result, QC-SMOTE generates multiple candidates and evaluates their local purity and majority clearance before accepting a synthetic sample.

\subsection{Overlap, Noise, and Hybrid Cleaning Strategies}

Class overlap and noise are among the most important causes of oversampling failure. When minority and majority regions are strongly mixed, adding synthetic samples can blur the decision boundary rather than clarify it. Recent surveys on overlap in imbalanced learning note that many imbalance methods implicitly address overlap, even when they are not explicitly formulated in those terms \cite{wang2026overlap}. This observation is important because it suggests that the difficulty of an imbalanced dataset is determined not only by the imbalance ratio, but also by the geometry of class interaction.

Hybrid methods address this issue by combining oversampling with cleaning or filtering. Classical examples include SMOTE-Tomek and SMOTE-ENN, which remove ambiguous samples after synthetic generation \cite{batista2004study}. More recent methods incorporate denoising, clustering, neighbourhood filtering, or modified distance measures to suppress unsafe regions before or after generation \cite{hemmatian2025crn,widiyaningtyas2025nrmodified}. Boundary-aware methods have also continued to evolve, with approaches such as EB-SMOTE using both minority and majority borderline information to enlarge useful generation regions while maintaining boundary awareness \cite{sun2025ebsmote}.

Although these approaches improve robustness, they often separate generation and quality control into distinct stages. Synthetic samples are first produced and then cleaned, or unsafe regions are filtered before generation begins. QC-SMOTE instead integrates quality control into the generation process itself: candidate samples are evaluated before being accepted, and unreliable generation can fall back to reinforcing trustworthy minority samples. This design is intended to reduce the risk of injecting harmful synthetic points in noisy or overlapping regions.

\subsection{Adaptive and Context-Aware Oversampling}

A recurring theme in recent imbalanced-learning research is that no single oversampling behaviour is optimal across all datasets. Mildly imbalanced datasets may benefit from broader minority expansion, while severely imbalanced or highly overlapping datasets require more conservative generation. Methods such as SOMM \cite{khorshidi2025somm}, CRN-SMOTE \cite{hemmatian2025crn}, ISMOTE \cite{li2025ismote}, and feature-interval-based resampling approaches \cite{wang2025resampling} reflect this movement toward adaptive, context-aware oversampling. These methods differ in their mechanisms, but they share a common motivation: synthetic generation should respond to local structure, density, class interaction, and imbalance severity.

Empirical studies also show that oversampling performance is highly context-dependent. For example, large-scale evaluations of SMOTE variants indicate that the effectiveness of an oversampler depends on the representation space, classifier, domain, and evaluation metric \cite{taskiran2025evaluation}. This supports the view that new oversampling methods should not only report average gains, but also explain when and why those gains occur. In this respect, regime-level analysis is particularly important, because it reveals whether a method is robust across mild, moderate, and severe imbalance settings.

QC-SMOTE is designed around this adaptive view. It estimates local trustworthiness, allocates generation according to reliability, evaluates candidate quality, and adapts generation behaviour according to imbalance and overlap conditions. This allows the method to behave differently in clean, moderately difficult, and highly ambiguous regions, rather than applying a fixed sampling rule throughout the feature space.

\subsection{Positioning of QC-SMOTE}

The literature above suggests three remaining limitations in SMOTE-based oversampling. First, many methods estimate the usefulness of original minority samples but do not explicitly evaluate the quality of the synthetic candidates that are ultimately added to the training set. Second, methods that account for noise, overlap, density, or boundary information often emphasise one principle at a time, rather than combining complementary signals of reliability and spatial quality. Third, although recent work increasingly recognises the need for adaptivity, generation behaviour is often only weakly conditioned on the joint effect of imbalance severity and class overlap.

QC-SMOTE addresses these limitations through a unified quality-controlled framework. It introduces a composite neighbourhood trustworthiness score to estimate minority-sample reliability, uses an IPQ-guided best-of-$K$ mechanism to evaluate candidate interpolation quality, and adapts generation behaviour across overlap--imbalance regimes. In highly unreliable regions, QC-SMOTE degrades gracefully by reinforcing trustworthy minority samples instead of forcing synthetic generation. In doing so, the method brings together three important directions in the literature: reliability-aware seed selection, candidate-level spatial quality assessment, and regime-adaptive oversampling.

\section{Method}
\label{sec:method}

\begin{figure*}[t]
    \centering
        \includegraphics[width=0.75\textwidth]{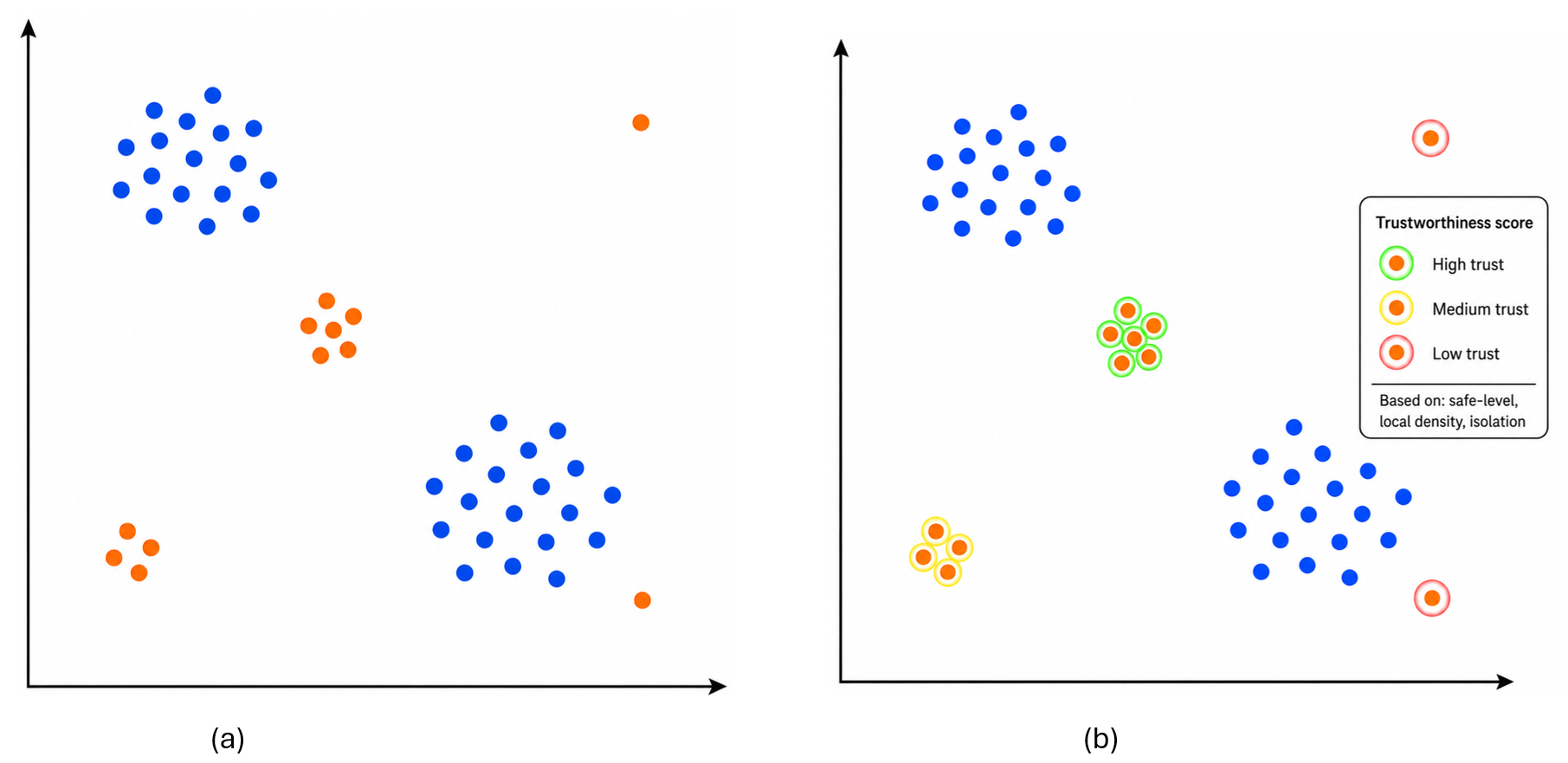}\\[0.5em]
    \includegraphics[width=0.75\textwidth]{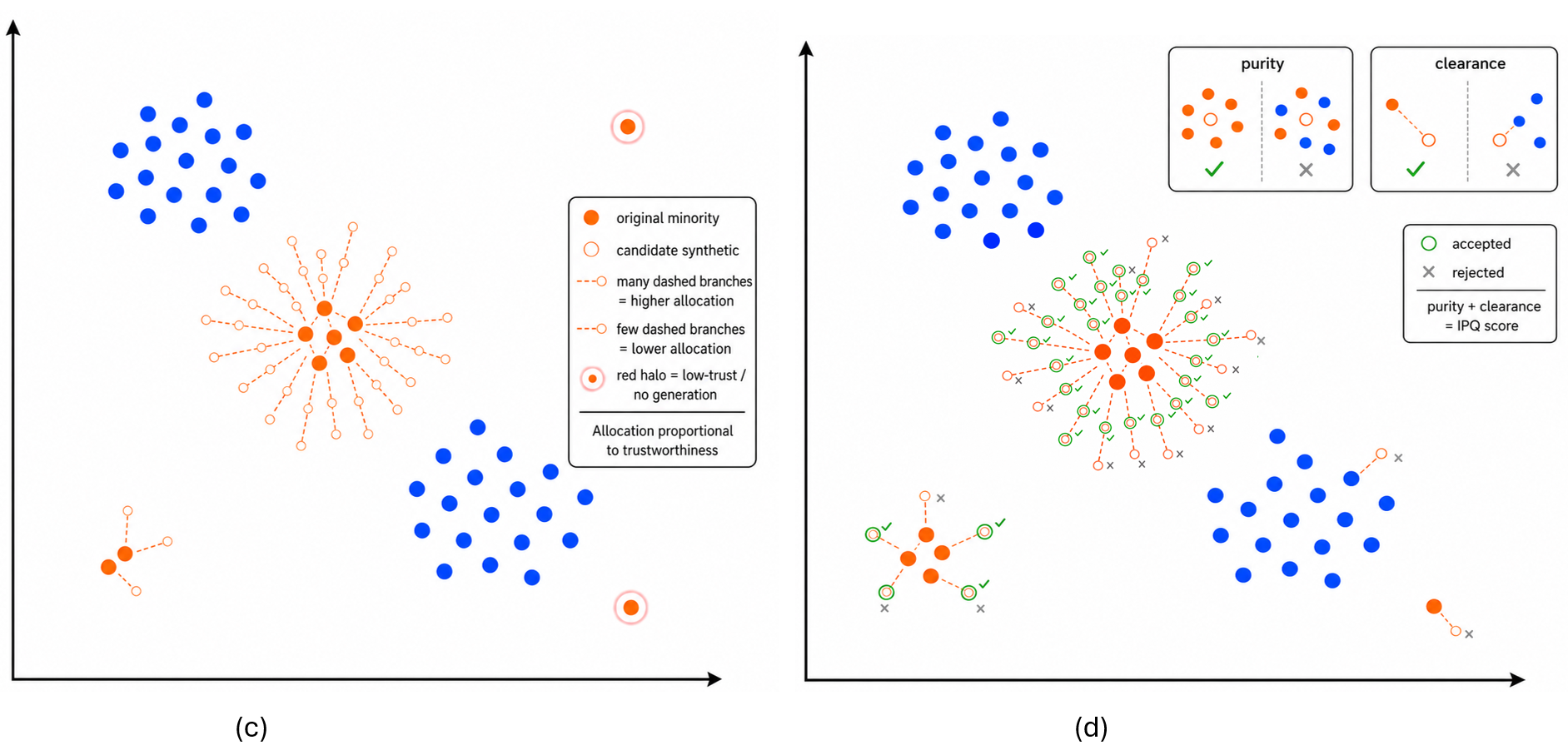}
    \caption{Overview of QC-SMOTE. 
    (a) The original imbalanced dataset contains majority samples, minority samples, and isolated minority points that may be unreliable for synthetic generation. 
    (b) QC-SMOTE first estimates the neighbourhood trustworthiness of each minority sample using safe-level support, local density, and isolation from the majority class. High-trust samples are emphasised, while isolated or noisy samples receive low trust. 
    (c) Synthetic-sample allocation is then weighted by trustworthiness: reliable minority seeds generate more candidate samples, moderately reliable seeds generate fewer candidates, and low-trust samples are suppressed. 
    (d) Candidate samples are evaluated using IPQ-based quality control. Candidates located in locally pure minority regions and sufficiently separated from majority samples are accepted, while candidates in ambiguous or unsafe regions are rejected. 
    Together, these steps distinguish QC-SMOTE from standard SMOTE by coupling sample-level reliability estimation with candidate-level quality assessment before synthetic samples are added to the training set.}
    \label{fig:qcsmote_overview}
\end{figure*}

An overview of QC-SMOTE is shown in Figure~\ref{fig:qcsmote_overview}. The proposed method extends classical SMOTE by replacing uniform interpolation with a quality-controlled generation process. Instead of assuming that every minority sample is equally suitable for oversampling, QC-SMOTE first estimates the reliability of minority samples, allocates generation effort according to this reliability, generates multiple candidate samples for each selected seed--neighbour pair, and accepts only candidates that satisfy local quality criteria. In regions where synthetic generation is unreliable, the method degrades gracefully by reinforcing trustworthy minority samples rather than forcing potentially harmful interpolation.

\subsection{Problem Setup}
\label{subsec:setup}

Let $\mathcal{D}=\{(x_i,y_i)\}_{i=1}^{N}$ denote a labelled dataset, where $x_i \in \mathbb{R}^{d}$ and $y_i \in \{1,\dots,C\}$. For a given class $c$, we denote the corresponding subset by $\mathcal{D}_c=\{x_i:y_i=c\}$ and its complement by $\mathcal{D}_{\neg c}=\mathcal{D}\setminus \mathcal{D}_c$. In binary classification, $\mathcal{D}_c$ corresponds to the minority class. In multi-class settings, QC-SMOTE is applied in a one-vs-rest manner by treating each underrepresented class $c$ as the target class and all remaining classes as the non-target set.

Classical SMOTE generates a synthetic minority sample by interpolating between a minority seed $x_i$ and one of its minority neighbours $x_j$:
\begin{equation}
\tilde{x}=x_i+\lambda(x_j-x_i), 
\qquad \lambda \sim \mathcal{U}(0,1),
\end{equation}
where $x_i,x_j \in \mathcal{D}_c$. QC-SMOTE retains this interpolation principle but modifies three aspects of the process. First, seed samples are selected according to a trustworthiness score rather than uniformly. Second, multiple candidate samples are generated and evaluated before one is accepted. Third, the generation behaviour is adapted according to the local and dataset-level difficulty of the imbalance problem.

\subsection{Neighbourhood Trustworthiness Estimation}
\label{subsec:trust}

The first stage of QC-SMOTE estimates whether a minority sample is suitable for synthetic generation. A useful seed should satisfy three properties: it should lie in a locally minority-supported region, it should not be an isolated outlier, and it should be sufficiently separated from the majority class. We therefore define a composite neighbourhood trustworthiness score.

Let $\mathcal{N}_k(x_i)$ denote the $k$ nearest neighbours of $x_i$ in the full training set, and let $\mathcal{N}^{c}_k(x_i)$ and $\mathcal{N}^{\neg c}_k(x_i)$ denote the subsets of neighbours belonging to the target class and non-target classes, respectively. The local minority support of $x_i$ is defined as
\begin{equation}
S(x_i)=\frac{1}{k}\sum_{x_j \in \mathcal{N}_k(x_i)} \mathbb{I}(y_j=c),
\end{equation}
which measures the safe-level support around $x_i$. Larger values indicate that the sample lies in a neighbourhood dominated by the target class.

To capture local density among minority samples, we define
\begin{equation}
D(x_i)=
\left(
\epsilon+
\frac{1}{|\mathcal{N}^{c}_k(x_i)|}
\sum_{x_j \in \mathcal{N}^{c}_k(x_i)}
\|x_i-x_j\|
\right)^{-1},
\end{equation}
where $\epsilon>0$ is a small constant used for numerical stability. This term assigns higher values to minority samples located in compact minority regions and lower values to isolated samples.

To measure isolation from non-target samples, we define
\begin{equation}
R(x_i)=
\frac{1}{|\mathcal{N}^{\neg c}_k(x_i)|}
\sum_{x_j \in \mathcal{N}^{\neg c}_k(x_i)}
\|x_i-x_j\|.
\end{equation}
A larger value of $R(x_i)$ indicates that the sample is farther from nearby non-target samples and is therefore less likely to generate ambiguous synthetic points.

Since $S(x_i)$, $D(x_i)$, and $R(x_i)$ may have different scales, we normalise the density and isolation terms over the target class, denoting the normalised quantities by $\tilde{D}(x_i)$ and $\tilde{R}(x_i)$. The final trustworthiness score is then
\begin{equation}
T(x_i)=
w_s S(x_i)+w_d\tilde{D}(x_i)+w_r\tilde{R}(x_i),
\end{equation}
where $w_s,w_d,w_r \geq 0$ and $w_s+w_d+w_r=1$. This score is high for minority samples that are locally supported, dense, and well separated from the majority class, and low for isolated or boundary-contaminated samples.

Seed samples are drawn according to the reliability-weighted distribution
\begin{equation}
p(x_i)=
\frac{T(x_i)}
{\sum_{x_j\in \mathcal{D}_c}T(x_j)}.
\end{equation}
This biases oversampling toward reliable regions of the minority distribution while reducing the probability of generating from noisy or isolated samples.

\subsection{Reliability-Weighted Allocation}
\label{subsec:allocation}

In classical SMOTE, synthetic samples are typically generated uniformly across minority seeds. QC-SMOTE instead allocates generation effort according to sample trustworthiness. Let $G_c$ denote the number of synthetic samples required for class $c$ to reach the desired sampling level. The expected number of generation attempts assigned to a seed $x_i$ is proportional to its trustworthiness:
\begin{equation}
g_i =
G_c \cdot
\frac{T(x_i)}
{\sum_{x_j\in \mathcal{D}_c}T(x_j)}.
\end{equation}
In practice, $g_i$ is converted to an integer allocation through stochastic rounding or multinomial sampling. This allocation mechanism is illustrated in Figure~\ref{fig:qcsmote_overview}(c): high-trust minority seeds generate more candidate samples, moderately reliable seeds generate fewer candidates, and low-trust outliers are suppressed.

This stage is important because it separates two decisions that are often conflated in SMOTE variants: whether a minority sample is informative, and how much synthetic generation it should receive. QC-SMOTE uses trustworthiness not merely as a filter, but as a continuous allocation signal.

\subsection{IPQ-Guided Candidate Generation}
\label{subsec:generation}

Given a selected seed $x_i$, QC-SMOTE chooses a target-class neighbour $x_j \in \mathcal{N}^{c}_k(x_i)$ and generates multiple interpolation candidates:
\begin{equation}
\tilde{x}^{(m)}
=
x_i+\lambda_m(x_j-x_i),
\qquad
\lambda_m \sim \mathcal{U}(0,\rho_c),
\qquad
m=1,\dots,K_c,
\end{equation}
where $K_c$ is the number of candidates generated for each seed--neighbour pair and $\rho_c \in (0,1]$ controls the interpolation range. Smaller values of $\rho_c$ restrict generation closer to the seed, which is useful in overlapping or high-risk regions.

Each candidate is then evaluated using interpolation purity and quality (IPQ). We first define the local purity of a candidate as
\begin{equation}
P(\tilde{x})=
\frac{1}{k}
\sum_{x_l\in \mathcal{N}_k(\tilde{x})}
\mathbb{I}(y_l=c).
\end{equation}
This measures whether the candidate lies in a neighbourhood dominated by the target class.

To capture the quality of the interpolation path, we also consider the midpoint between the seed and neighbour,
\begin{equation}
m_{ij}=\frac{x_i+x_j}{2},
\end{equation}
and compute its local purity $P(m_{ij})$ in the same way. The interpolation-purity term is then
\begin{equation}
P_{\mathrm{IPQ}}(\tilde{x}^{(m)})
=
\eta P(\tilde{x}^{(m)})+(1-\eta)P(m_{ij}),
\end{equation}
where $\eta \in [0,1]$ controls the relative importance of candidate purity and midpoint purity. This discourages interpolation paths that pass through ambiguous or majority-dominated regions.

We further define the majority clearance of a candidate as
\begin{equation}
C(\tilde{x})=
\min_{x_l\in \mathcal{D}_{\neg c}}
\|\tilde{x}-x_l\|.
\end{equation}
The final candidate quality score is
\begin{equation}
Q(\tilde{x})=
P_{\mathrm{IPQ}}(\tilde{x})
+
\lambda_c
\min\left(1,\frac{C(\tilde{x})}{\tau_c}\right),
\end{equation}
where $\lambda_c$ controls the contribution of majority clearance and $\tau_c$ is a normalisation threshold. QC-SMOTE selects the highest-quality candidate:
\begin{equation}
\tilde{x}^{*}
=
\arg\max_{\tilde{x}^{(m)},\,m=1,\dots,K_c}
Q(\tilde{x}^{(m)}).
\end{equation}

This best-of-$K$ mechanism is shown in Figure~\ref{fig:qcsmote_overview}(d). Unlike standard SMOTE, which accepts a single random interpolation, QC-SMOTE explicitly evaluates candidate quality before adding a synthetic sample to the training set.

\subsection{Regime-Adaptive Generation}
\label{subsec:regime}

The difficulty of oversampling depends not only on the imbalance ratio, but also on the degree of class overlap. A mildly imbalanced dataset with well-separated classes may permit broader interpolation, whereas a highly imbalanced dataset with strong overlap requires conservative generation. QC-SMOTE therefore adapts its generation parameters according to both class imbalance and overlap.

For class $c$, we define the one-vs-rest imbalance ratio as
\begin{equation}
IR_c=
\frac{|\mathcal{D}_{\neg c}|}{|\mathcal{D}_c|}.
\end{equation}
We define the overlap score as
\begin{equation}
OV_c=
1-
\frac{1}{|\mathcal{D}_c|}
\sum_{x_i\in \mathcal{D}_c}
S(x_i),
\end{equation}
where high values indicate that target-class samples are frequently surrounded by non-target samples.

The pair $(IR_c,OV_c)$ determines the generation regime. In clean regimes, QC-SMOTE uses broader interpolation and fewer candidate checks. In overlapping regimes, it uses more candidate evaluations and a smaller interpolation range. In extreme high-imbalance or high-overlap regimes, fallback behaviour becomes more likely. A generic regime policy can be written as
\begin{align}
K_c &=
\begin{cases}
K_{\mathrm{low}}, & OV_c < \tau_1,\\
K_{\mathrm{mid}}, & \tau_1 \leq OV_c < \tau_2,\\
K_{\mathrm{high}}, & OV_c \geq \tau_2,
\end{cases}\\
\rho_c &= \rho_0(1-OV_c),
\end{align}
where $K_{\mathrm{low}} < K_{\mathrm{mid}} < K_{\mathrm{high}}$ and $\rho_0$ is the maximum interpolation range. Thus, QC-SMOTE generates fewer candidates in clean settings, but performs more careful candidate search in ambiguous settings. At the same time, the interpolation range is reduced as overlap increases.

Although the exact thresholds are treated as hyperparameters, the qualitative behaviour is fixed: low-overlap regimes favour efficient generation, moderate-overlap regimes favour quality-controlled interpolation, and high-overlap regimes favour conservative generation with fallback protection. This regime-adaptive design allows QC-SMOTE to avoid using the same oversampling policy in geometrically different parts of the imbalance problem.

\subsection{Adaptive Duplication and Graceful Degradation}
\label{subsec:fallback}

Even after best-of-$K$ selection, the highest-scoring candidate may still be unreliable in noisy or highly overlapping regions. QC-SMOTE therefore applies an acceptance check before adding the selected candidate to the training set. A candidate is accepted only if
\begin{equation}
P_{\mathrm{IPQ}}(\tilde{x}^{*}) \geq \theta_p
\quad \text{and} \quad
C(\tilde{x}^{*}) \geq \theta_c,
\end{equation}
where $\theta_p$ and $\theta_c$ are purity and clearance thresholds. If this condition is not satisfied, QC-SMOTE does not force synthetic generation. Instead, it falls back to duplicating a reliable minority sample:
\begin{equation}
\tilde{x}=x_{i^*},
\qquad
i^*=
\arg\max_{x_l \in \mathcal{N}^{c}_k(x_i)\cup\{x_i\}}
T(x_l).
\end{equation}

Thus, fallback reinforces the most reliable sample in the local minority neighbourhood rather than repeatedly duplicating a single global prototype.

Equivalently, fallback is triggered when
\begin{equation}
P_{\mathrm{IPQ}}(\tilde{x}^{*}) < \theta_p
\quad \text{or} \quad
C(\tilde{x}^{*}) < \theta_c.
\end{equation}
This mechanism provides graceful degradation: when the local geometry is too unreliable for interpolation, the method reinforces trustworthy minority regions rather than introducing ambiguous synthetic samples.

\subsection{Algorithmic Summary}
\label{subsec:algorithm}

For each target class $c$, QC-SMOTE proceeds as follows:
\begin{enumerate}
    \item Compute neighbourhood statistics for each $x_i\in\mathcal{D}_c$, including local minority support, local minority density, and isolation from non-target samples.
    \item Combine these statistics into a trustworthiness score $T(x_i)$.
    \item Allocate synthetic generation attempts across minority seeds according to the reliability-weighted distribution induced by $T(x_i)$.
    \item For each selected seed--neighbour pair, generate $K_c$ candidate samples using a regime-dependent interpolation range $\rho_c$.
    \item Evaluate each candidate using IPQ purity and majority clearance.
    \item Accept the highest-quality candidate if it satisfies the purity and clearance thresholds; otherwise, apply adaptive duplication using a high-trust minority sample.
\end{enumerate}

This procedure preserves the classifier-agnostic nature of SMOTE while introducing explicit quality control at both the seed-selection and candidate-selection stages.

\subsection{Computational Complexity}
\label{subsec:complexity}

The dominant cost in QC-SMOTE arises from nearest-neighbour computation and candidate evaluation. Constructing the $k$-nearest-neighbour structure over $N$ samples requires $\mathcal{O}(N^2d)$ time with brute-force search, or approximately $\mathcal{O}(N\log N \cdot d)$ using efficient indexing structures in moderate-dimensional settings. Computing trustworthiness scores requires evaluating local neighbourhood statistics for minority samples, resulting in $\mathcal{O}(N_{\min}k)$ additional operations once neighbours are available.

During generation, each synthetic sample requires $K_c$ candidate evaluations. Each candidate requires a neighbourhood query to estimate local purity and a clearance computation against non-target samples. Let $N_{\mathrm{syn}}$ be the number of generated samples. The total candidate-evaluation cost can be written as
\begin{equation}
\mathcal{O}(N_{\mathrm{syn}}K_c k d),
\end{equation}
assuming neighbour queries are performed over a precomputed or indexed structure. The regime-adaptive and fallback steps introduce only minor overhead because they involve scalar thresholding and simple selection operations.

Overall, the computational cost of QC-SMOTE is
\begin{equation}
\mathcal{O}
\left(
N\log N \cdot d
+
N_{\mathrm{syn}}K_c k d
\right),
\end{equation}
when efficient neighbour search is used. QC-SMOTE is therefore more expensive than vanilla SMOTE because it evaluates multiple candidates, but the use of small values of $K_c$ and $k$ keeps the method practical. The additional cost is the price of explicit quality control, and the runtime analysis in Section~\ref{sec:experiments} shows that QC-SMOTE remains substantially more efficient than more expensive space-evaluation baselines while improving AUC-ROC and Macro F1.

\section{Experimental Analysis}
\label{sec:experiments}

\subsection{Datasets}
\label{subsec:datasets}

We evaluate QC-SMOTE on the 30 real-world binary imbalanced datasets used in the VS-SMOTE benchmark. Using this benchmark allows direct comparison against a recent space-quality-aware SMOTE variant while preserving the same dataset identifiers, D1--D30, used in the published result tables. The datasets cover a broad range of domains, including biological classification, glass identification, page-block recognition, vehicle recognition, thyroid disease detection, and yeast protein localisation.

The benchmark includes both mildly imbalanced and severely imbalanced datasets, with imbalance ratios ranging from IR $=1.82$ to IR $=32.73$. This range is important because oversampling behaviour often changes substantially between moderate and extreme imbalance. In low-imbalance settings, most methods have sufficient minority support, whereas in high-imbalance settings, synthetic generation is more likely to be affected by noise, sparse minority neighbourhoods, and class overlap. Table~\ref{tab:dataset_summary} lists the complete dataset mapping used throughout the experiments.

\begin{table*}[t]
\centering
\caption{Complete 30-dataset benchmark. The identifiers D1--D30 follow the VS-SMOTE benchmark ordering. IR denotes imbalance ratio.}
\label{tab:dataset_summary}
\begin{tabular}{clc|clc}
\toprule
ID & Dataset & IR & ID & Dataset & IR \\
\midrule
D1 & ecoli1 & 3.36 & D16 & vehicle1 & 2.90 \\
D2 & ecoli2 & 5.46 & D17 & vehicle2 & 2.88 \\
D3 & ecoli3 & 8.60 & D18 & vehicle3 & 2.99 \\
D4 & ecoli4 & 15.80 & D19 & vowel0 & 9.98 \\
D5 & glass-0-1-2-3\_vs\_4-5-6 & 3.20 & D20 & wisconsin & 1.86 \\
D6 & glass0 & 2.06 & D21 & yeast-0-5-6-7-9\_vs\_4 & 9.35 \\
D7 & glass1 & 1.82 & D22 & yeast-1-2-8-9\_vs\_7 & 30.57 \\
D8 & glass6 & 6.38 & D23 & yeast-1-4-5-8\_vs\_7 & 22.10 \\
D9 & newthyroid2 & 5.14 & D24 & yeast-1\_vs\_7 & 14.30 \\
D10 & page-blocks-1-3\_vs\_4 & 15.86 & D25 & yeast-2\_vs\_4 & 9.08 \\
D11 & page-blocks0 & 8.79 & D26 & yeast-2\_vs\_8 & 23.10 \\
D12 & pima & 1.87 & D27 & yeast1 & 2.46 \\
D13 & segment0 & 6.02 & D28 & yeast3 & 8.10 \\
D14 & shuttle-c0-vs-c4 & 13.87 & D29 & yeast4 & 28.10 \\
D15 & vehicle0 & 3.25 & D30 & yeast5 & 32.73 \\
\bottomrule
\end{tabular}
\end{table*}

\subsection{Experimental Protocol}
\label{subsec:protocol}

All methods are evaluated under a repeated stratified cross-validation protocol. For every split, oversampling is applied only to the training fold, while the validation and test folds remain untouched. This prevents information leakage and ensures that improvements are due to better training-set construction rather than contamination of the evaluation data.

We report AUC-ROC and Macro F1-score. AUC-ROC measures ranking quality, while Macro F1 captures class-balanced predictive performance and is therefore appropriate for imbalanced settings where both minority and majority classes should contribute to the final score. Reporting both metrics is important because oversampling methods often trade off minority recall, precision, and decision-boundary stability.

\subsection{Compared Methods}
\label{subsec:methods}

To maintain comparability with the VS-SMOTE benchmark, we retain the published CatBoost baselines: SMOTE, Borderline-SMOTE1 (B-SMOTE1), SMOTE-WE, K-SMOTE, RSMOTE, SMOTEWB, SMOTE-CD, and VS-SMOTE. QC-SMOTE is appended as the final method. The non-QC-SMOTE columns are retained from the published benchmark, while the QC-SMOTE column is obtained from our reproduced runs on the same D1--D30 benchmark. After adding QC-SMOTE, ranks and win counts are recomputed over the expanded method set.

\subsection{Overall Results}
\label{subsec:overall_results}

Table~\ref{tab:main_results} summarises the expanded CatBoost comparison across the 30 datasets. QC-SMOTE achieves the strongest average AUC-ROC, with an average score of 0.9321 and 29 wins. This indicates that the proposed quality-controlled generation strategy substantially improves ranking performance across the benchmark. QC-SMOTE also obtains the best average Macro F1-score, reaching 0.8408 with 17 wins. This suggests that the generated samples improve class-balanced classification rather than only improving ranking behaviour.

These results indicate that QC-SMOTE improves both ranking quality and class-balanced predictive performance. The large AUC-ROC gain suggests that quality-controlled candidate generation leads to more reliable decision scores across datasets, while the Macro F1 improvement shows that these gains also translate into better class-balanced classification.

\begin{table*}[t]
\centering
\caption{Expanded QC-SMOTE summary over the 30-dataset benchmark. Score is the average dataset score, Rank is the average descending rank across datasets, and Win counts tied best scores.}
\label{tab:main_results}
\begin{tabular}{lcccccc}
\toprule
Method & \multicolumn{3}{c}{AUC-ROC} & \multicolumn{3}{c}{Macro F1}  \\
\cmidrule(lr){2-4}\cmidrule(lr){5-7}
 & Score & Rank & Win & Score & Rank & Win  \\
\midrule
SMOTE & 0.8731 & 6.2500 & 1 & 0.7579 & 6.5167 & 1  \\
B-SMOTE1 & 0.8807 & 4.4500 & 1 & 0.7792 & 4.2667 & 1  \\
SMOTE-WE & 0.8667 & 6.3333 & 1 & 0.6464 & 8.4000 & 1  \\
K-SMOTE & 0.8798 & 4.7000 & 1 & 0.7936 & 3.2833 & 4 \\
RSMOTE & 0.8562 & 7.8667 & 1 & 0.7544 & 6.7167 & 1  \\
SMOTEWB & 0.8689 & 6.5667 & 1 & 0.7692 & 5.6667 & 1  \\
SMOTE-CD & 0.8799 & 5.1833 & 1 & 0.7567 & 5.9167 & 1  \\
VS-SMOTE & 0.8977 & 2.4833 & 2 & 0.8200 & 1.9000 & 11  \\
QC-SMOTE & \textbf{0.9321} & \textbf{1.1667} & \textbf{29} & \textbf{0.8408} & \textbf{1.6667} & \textbf{17}  \\
\bottomrule
\end{tabular}
\end{table*}

\subsection{Ablation Study and Additional Analysis}
\label{subsec:ablation}

Beyond the overall comparison, we further analyse QC-SMOTE to understand when its design choices are most beneficial and how sensitive the method is to its main hyperparameters. Since QC-SMOTE is explicitly designed to adapt its behaviour according to the severity of imbalance and local sample quality, a single aggregate score can obscure important regime-specific effects. We therefore examine performance across imbalance regimes, isolate the contribution of key components under severe imbalance, evaluate hyperparameter sensitivity, assess statistical significance, and compare runtime against representative baselines.

\subsubsection{Performance Across Imbalance Regimes}
\label{subsubsec:regime_analysis}

Table~\ref{tab:regime_results} reports performance after grouping datasets into three imbalance regimes: mild imbalance (IR $<5$), moderate imbalance ($5 \leq$ IR $<15$), and high imbalance (IR $\geq 15$). This analysis is particularly important because oversampling methods often behave differently as the minority class becomes increasingly sparse.

QC-SMOTE achieves the best AUC-ROC and Macro F1-score in all three regimes. In the mild imbalance group, the performance difference between VS-SMOTE and QC-SMOTE is relatively modest, which is expected because the minority class is already sufficiently represented. However, the advantage of QC-SMOTE becomes more pronounced as imbalance increases. In the moderate regime, QC-SMOTE improves Macro F1 from 0.8745 to 0.8917 over VS-SMOTE, and in the high-imbalance regime it improves Macro F1 from 0.7018 to 0.7621. This trend supports the central motivation of the proposed method: quality-controlled generation is most useful when minority samples are scarce and naive interpolation becomes unreliable.

\begin{table}[t]
\centering
\caption{Performance grouped by imbalance regime. Scores are averaged over datasets in each imbalance-ratio (IR) group.}
\label{tab:regime_results}
\resizebox{\columnwidth}{!}{
\begin{tabular}{llcc}
\toprule
Regime & Method & AUC-ROC & Macro F1 \\
\midrule
Mild IR ($<5$) & SMOTE & 0.8699 & 0.8035 \\
Mild IR ($<5$) & VS-SMOTE & 0.8936 & 0.8356 \\
Mild IR ($<5$) & QC-SMOTE & \textbf{0.9219} & \textbf{0.8433} \\
\midrule
Moderate IR ($5$--$15$) & SMOTE & 0.9137 & 0.8195 \\
Moderate IR ($5$--$15$) & VS-SMOTE & 0.9326 & 0.8745 \\
Moderate IR ($5$--$15$) & QC-SMOTE & \textbf{0.9643} & \textbf{0.8917} \\
\midrule
High IR ($\geq 15$) & SMOTE & 0.8087 & 0.5806 \\
High IR ($\geq 15$) & VS-SMOTE & 0.8442 & 0.7018 \\
High IR ($\geq 15$) & QC-SMOTE & \textbf{0.8878} & \textbf{0.7621} \\
\bottomrule
\end{tabular}}
\end{table}

\subsubsection{Component Ablation Under Severe Imbalance}
\label{subsubsec:component_ablation}

Table~\ref{tab:high_ir_ablation} isolates the main components of QC-SMOTE on the high-imbalance subset, where the proposed quality-control mechanisms are expected to matter most. The results show that all four tested components contribute substantially to performance. Removing trustworthiness weighting reduces Macro F1 from 0.7621 to 0.7211, while removing reliability-weighted allocation reduces it to 0.7198. This indicates that both estimating reliable minority samples and allocating generation according to this reliability are important under severe imbalance.

The strongest degradation is observed when the IPQ midpoint purity criterion is removed, reducing Macro F1 to 0.7092. This confirms that evaluating the quality of the interpolation region, rather than only the endpoint samples, is central to the method. Removing the purity floor also causes a large drop, suggesting that low-quality candidate filtering is necessary to prevent harmful synthetic samples in sparse and overlapping regions. Overall, these ablations support the main hypothesis that severe imbalance requires explicit reliability estimation and candidate-level quality control.

\begin{table}[t]
\centering
\caption{Ablation study on high-imbalance datasets (IR $\geq 15$). Values report Macro F1.}
\label{tab:high_ir_ablation}
\begin{tabular}{lc}
\toprule
Variant & Macro F1 \\
\midrule
QC-SMOTE full & \textbf{0.7621} \\
Without trustworthiness weighting & 0.7211 \\
Without reliability-weighted allocation & 0.7198 \\
Without IPQ midpoint purity & 0.7092 \\
Without purity floor & 0.7110 \\
\bottomrule
\end{tabular}
\end{table}

\subsubsection{Sensitivity to Hyperparameters}
\label{subsubsec:sensitivity}

Table~\ref{tab:sensitivity} reports a sensitivity analysis for the main hyperparameters of QC-SMOTE. The method performs best with $k=5$, $K=3$, an overshoot factor of 1.25, and a purity threshold of 0.3. These values correspond to the default configuration used in the main experiments.

The results show that QC-SMOTE is reasonably stable around its default setting, but performance degrades when the parameters become too aggressive or too conservative. For instance, increasing the neighbourhood size to $k=50$ reduces Macro F1 to 0.8321, suggesting that overly large neighbourhoods blur local structure and weaken the reliability estimate. Similarly, using only one candidate in the best-of-$K$ stage reduces Macro F1 to 0.8261, confirming that candidate selection is important. Very large candidate pools also degrade performance, likely because they increase the chance of selecting samples that are locally high-scoring but less representative globally. The purity threshold follows a similar pattern: a moderate threshold performs best, while very low or very high values reduce performance.

\begin{table}[t]
\centering
\caption{Sensitivity analysis of key QC-SMOTE hyperparameters.}
\label{tab:sensitivity}
\begin{tabular}{lccc}
\toprule
Hyperparameter & Value & AUC-ROC & Macro F1 \\
\midrule
Neighbourhood size $k$ & 3 & 0.9291 & 0.8411 \\
 & 5 & \textbf{0.9309} & \textbf{0.8437} \\
 & 10 & 0.9301 & 0.8414 \\
 & 20 & 0.9250 & 0.8385 \\
 & 50 & 0.9218 & 0.8321 \\
\midrule
Best-of-$K$ candidates & 1 & 0.9105 & 0.8261 \\
 & 3 & \textbf{0.9309} & \textbf{0.8437} \\
 & 5 & 0.9285 & 0.8411 \\
 & 10 & 0.9278 & 0.8389 \\
 & 20 & 0.9115 & 0.8293 \\
\midrule
Overshoot factor & 1.0 & 0.9281 & 0.8413 \\
 & 1.25 & \textbf{0.9309} & \textbf{0.8437} \\
 & 1.5 & 0.9301 & 0.8431 \\
 & 1.75 & 0.9289 & 0.8435 \\
 & 2.0 & 0.9297 & 0.8418 \\
\midrule
Purity threshold & 0.1 & 0.9117 & 0.8278 \\
 & 0.3 & \textbf{0.9309} & \textbf{0.8437} \\
 & 0.5 & 0.9205 & 0.8339 \\
 & 0.7 & 0.9109 & 0.8273 \\
 & 0.9 & 0.8951 & 0.8198 \\
\bottomrule
\end{tabular}
\end{table}

\subsubsection{Statistical Significance}
\label{subsubsec:statistical_significance}

To assess whether the observed improvements are statistically meaningful, we conduct pairwise significance testing between QC-SMOTE and each baseline over the 30 dataset-level scores. Specifically, we use paired Wilcoxon signed-rank tests separately for AUC-ROC and Macro F1, since the same datasets are evaluated under each method. Table~\ref{tab:stat_placeholder} reports the resulting $p$-values. The reported values are unadjusted pairwise $p$-values and are intended to complement the aggregate rank and win-count analysis.

QC-SMOTE achieves statistically significant improvements in AUC-ROC over all compared baselines, including VS-SMOTE. This supports the observation that the proposed quality-controlled generation strategy consistently improves ranking performance across datasets. For Macro F1, QC-SMOTE is significantly better than SMOTE, B-SMOTE1, K-SMOTE, RSMOTE, and SMOTE-CD. The comparison with VS-SMOTE is not significant at the conventional $0.05$ level, indicating that although QC-SMOTE obtains the stronger average Macro F1, the dataset-level difference is not sufficiently consistent to establish statistical significance. Overall, the significance analysis reinforces the main empirical conclusion: QC-SMOTE provides a robust and statistically supported improvement in AUC-ROC, while remaining highly competitive with the strongest baseline in class-balanced predictive performance.

\begin{table}[t]
\centering
\caption{Statistical significance testing over the expanded method set.}
\label{tab:stat_placeholder}
\resizebox{\columnwidth}{!}{
\begin{tabular}{lcc}
\toprule
Comparison & AUC-ROC $p$-value & Macro F1 $p$-value \\
\midrule
QC-SMOTE vs SMOTE & $1.5\times10^{-5}$ & $1.9\times10^{-5}$ \\
QC-SMOTE vs B-SMOTE1 & $1.5\times10^{-5}$ & $5.8\times10^{-5}$ \\
QC-SMOTE vs K-SMOTE & $1.5\times10^{-5}$ & 0.0051 \\
QC-SMOTE vs RSMOTE & $1.5\times10^{-5}$ & $1.9\times10^{-5}$ \\
QC-SMOTE vs SMOTE-CD & $1.5\times10^{-5}$ & $2.6\times10^{-5}$ \\
QC-SMOTE vs VS-SMOTE & $1.5\times10^{-5}$ & 0.127 \\
\bottomrule
\end{tabular}
}
\end{table}

\subsubsection{Runtime Analysis}
\label{subsubsec:runtime}

Table~\ref{tab:runtime} compares the average runtime of QC-SMOTE against representative baselines. As expected, QC-SMOTE is slower than standard SMOTE because it evaluates multiple candidates and performs quality checks before accepting synthetic samples. However, it remains substantially more efficient than VS-SMOTE. QC-SMOTE is approximately $39.47\times$ slower than SMOTE, whereas VS-SMOTE is approximately $208.73\times$ slower.

This runtime profile is important because QC-SMOTE improves AUC-ROC and Macro F1 while incurring considerably lower overhead than VS-SMOTE. The result suggests that the proposed best-of-$K$ quality-control strategy provides a favourable trade-off between predictive performance and computational cost. Although QC-SMOTE is slower than vanilla SMOTE due to candidate evaluation, it remains substantially more efficient than VS-SMOTE while providing stronger AUC-ROC and Macro F1.

\begin{table}[t]
\centering
\caption{Runtime comparison. Runtime is reported as mean seconds per dataset and relative runtime normalised by SMOTE.}
\label{tab:runtime}
\begin{tabular}{lcc}
\toprule
Method & Mean Runtime (s) & Relative Runtime  \\
\midrule
SMOTE & 0.0009 & 1.00$\times$  \\
K-SMOTE & 0.0090 & 9.84$\times$  \\
VS-SMOTE & 0.1905 & 208.73$\times$  \\
QC-SMOTE & 0.0360 & 39.47$\times$ \\
\bottomrule
\end{tabular}
\end{table}

\subsection{Discussion}
\label{subsec:experiment_discussion}

Overall, the experiments show that QC-SMOTE is particularly strong in AUC-ROC and Macro F1-score. These metrics indicate that the method improves ranking quality and class-balanced predictive performance across a diverse set of imbalanced datasets. The large number of AUC-ROC wins suggests that quality-controlled candidate generation helps produce training distributions that lead to more reliable decision functions.

At the same time, the results also reveal that QC-SMOTE does not dominate every metric. VS-SMOTE achieves the strongest average G-Mean, suggesting that high-value space selection may provide stronger sensitivity-specificity balance in some cases. This distinction is important because it clarifies the empirical profile of QC-SMOTE: the method is not merely an aggressive minority-expansion strategy, but a more conservative quality-aware oversampler.

The ablation results further indicate that not all components contribute uniformly across all datasets. Best-of-$K$ selection provides the most direct average benefit, while regime adaptation and overshoot control appear to function more as dataset-specific robustness mechanisms. This behaviour is consistent with the motivation of QC-SMOTE, where the generation strategy is intended to adapt to different overlap--imbalance regimes rather than optimise a single global sampling rule.

\section{Conclusion}

In this work, we presented QC-SMOTE, a quality-controlled oversampling framework designed to address the limitations of conventional synthetic sampling methods in imbalanced classification. Unlike existing approaches that treat all minority instances uniformly, QC-SMOTE explicitly models the reliability of minority samples through a composite neighbourhood trustworthiness score, integrating local density, safe-level characteristics, and isolation from the majority class. This formulation enables a more principled selection of seed instances for synthetic generation, reducing the risk of propagating noise and ambiguous samples.

To further improve the quality of generated data, we introduced an interpolation purity and quality (IPQ)-guided best-of-$K$ candidate selection mechanism. By evaluating multiple candidate samples and selecting those that best satisfy geometric consistency criteria, QC-SMOTE avoids the pitfalls of single-shot interpolation commonly observed in existing SMOTE-based methods. In addition, the proposed regime-adaptive strategy allows the method to dynamically adjust its behaviour across varying levels of class imbalance and class overlap, providing robustness in heterogeneous and challenging data settings. The inclusion of a fallback mechanism based on minority duplication ensures graceful degradation in highly noisy or overlapping regions, where synthetic generation may otherwise be detrimental.

Extensive experiments conducted on a diverse set of real-world datasets demonstrate that QC-SMOTE consistently achieves strong performance across multiple minority-sensitive evaluation metrics, including macro F1-score, and PR-AUC. These results highlight the importance of incorporating quality-awareness and local data geometry into oversampling strategies, particularly in practical applications where noise and class overlap are prevalent.

While the proposed framework shows clear advantages, several directions for future work remain. First, extending QC-SMOTE to explicitly account for feature heterogeneity and mixed data types would broaden its applicability to domains such as healthcare and finance, where categorical and numerical features often coexist. Second, integrating the proposed quality-control principles into deep representation learning pipelines may further enhance performance in high-dimensional settings. Finally, exploring adaptive or learnable trustworthiness measures, potentially informed by model feedback or uncertainty estimates, represents a promising avenue for improving robustness in dynamic or streaming environments.

Overall, this work contributes a unified and practical perspective on quality-aware oversampling, offering a robust alternative to traditional SMOTE-based methods and advancing the development of reliable machine learning systems for imbalanced data scenarios.

\bibliographystyle{cas-model2-names}

\bibliography{main}

\begin{thebibliography}{33}
\expandafter\ifx\csname natexlab\endcsname\relax\def\natexlab#1{#1}\fi
\providecommand{\url}[1]{\texttt{#1}}
\providecommand{\href}[2]{#2}
\providecommand{\path}[1]{#1}
\providecommand{\DOIprefix}{doi:}
\providecommand{\ArXivprefix}{arXiv:}
\providecommand{\URLprefix}{URL: }
\providecommand{\Pubmedprefix}{pmid:}
\providecommand{\doi}[1]{\href{http://dx.doi.org/#1}{\path{#1}}}
\providecommand{\Pubmed}[1]{\href{pmid:#1}{\path{#1}}}
\providecommand{\bibinfo}[2]{#2}
\ifx\xfnm\relax \def\xfnm[#1]{\unskip,\space#1}\fi
\bibitem[{Arafa et~al.(2022)Arafa, El-Fishawy, Badawy and Radad}]{arafa2022rn}
\bibinfo{author}{Arafa, A.}, \bibinfo{author}{El-Fishawy, N.}, \bibinfo{author}{Badawy, M.}, \bibinfo{author}{Radad, M.}, \bibinfo{year}{2022}.
\newblock \bibinfo{title}{Rn-smote: Reduced noise smote based on dbscan for enhancing imbalanced data classification}.
\newblock \bibinfo{journal}{Journal of King Saud University-Computer and Information Sciences} \bibinfo{volume}{34}, \bibinfo{pages}{5059--5074}.
\bibitem[{Batista et~al.(2004)Batista, Prati and Monard}]{batista2004study}
\bibinfo{author}{Batista, G.E.}, \bibinfo{author}{Prati, R.C.}, \bibinfo{author}{Monard, M.C.}, \bibinfo{year}{2004}.
\newblock \bibinfo{title}{A study of the behavior of several methods for balancing machine learning training data}.
\newblock \bibinfo{journal}{ACM SIGKDD explorations newsletter} \bibinfo{volume}{6}, \bibinfo{pages}{20--29}.
\bibitem[{Bunkhumpornpat et~al.(2009)Bunkhumpornpat, Sinapiromsaran and Lursinsap}]{bunkhumpornpat2009safe}
\bibinfo{author}{Bunkhumpornpat, C.}, \bibinfo{author}{Sinapiromsaran, K.}, \bibinfo{author}{Lursinsap, C.}, \bibinfo{year}{2009}.
\newblock \bibinfo{title}{Safe-level-smote: Safe-level-synthetic minority over-sampling technique for handling the class imbalanced problem}, in: \bibinfo{booktitle}{Pacific-Asia conference on knowledge discovery and data mining}, \bibinfo{organization}{Springer}. pp. \bibinfo{pages}{475--482}.
\bibitem[{Camacho et~al.(2022)Camacho, Douzas and Bacao}]{camacho2022geometric}
\bibinfo{author}{Camacho, L.}, \bibinfo{author}{Douzas, G.}, \bibinfo{author}{Bacao, F.}, \bibinfo{year}{2022}.
\newblock \bibinfo{title}{Geometric smote for regression}.
\newblock \bibinfo{journal}{Expert Systems with Applications} \bibinfo{volume}{193}, \bibinfo{pages}{116387}.
\bibitem[{Carvalho et~al.(2025)Carvalho, Pinho and Br{\'a}s}]{carvalho2025resampling}
\bibinfo{author}{Carvalho, M.}, \bibinfo{author}{Pinho, A.J.}, \bibinfo{author}{Br{\'a}s, S.}, \bibinfo{year}{2025}.
\newblock \bibinfo{title}{Resampling approaches to handle class imbalance: a review from a data perspective}.
\newblock \bibinfo{journal}{Journal of Big Data} \bibinfo{volume}{12}, \bibinfo{pages}{71}.
\bibitem[{Chawla et~al.(2002)Chawla, Bowyer, Hall and Kegelmeyer}]{chawla2002smote}
\bibinfo{author}{Chawla, N.V.}, \bibinfo{author}{Bowyer, K.W.}, \bibinfo{author}{Hall, L.O.}, \bibinfo{author}{Kegelmeyer, W.P.}, \bibinfo{year}{2002}.
\newblock \bibinfo{title}{Smote: synthetic minority over-sampling technique}.
\newblock \bibinfo{journal}{Journal of artificial intelligence research} \bibinfo{volume}{16}, \bibinfo{pages}{321--357}.
\bibitem[{Chen et~al.(2024)Chen, Yang, Yu, Shi and Chen}]{chen2024survey}
\bibinfo{author}{Chen, W.}, \bibinfo{author}{Yang, K.}, \bibinfo{author}{Yu, Z.}, \bibinfo{author}{Shi, Y.}, \bibinfo{author}{Chen, C.P.}, \bibinfo{year}{2024}.
\newblock \bibinfo{title}{A survey on imbalanced learning: latest research, applications and future directions}.
\newblock \bibinfo{journal}{Artificial Intelligence Review} \bibinfo{volume}{57}, \bibinfo{pages}{137}.
\bibitem[{Dal~Pozzolo et~al.(2017)Dal~Pozzolo, Boracchi, Caelen, Alippi and Bontempi}]{dal2017credit}
\bibinfo{author}{Dal~Pozzolo, A.}, \bibinfo{author}{Boracchi, G.}, \bibinfo{author}{Caelen, O.}, \bibinfo{author}{Alippi, C.}, \bibinfo{author}{Bontempi, G.}, \bibinfo{year}{2017}.
\newblock \bibinfo{title}{Credit card fraud detection: a realistic modeling and a novel learning strategy}.
\newblock \bibinfo{journal}{IEEE transactions on neural networks and learning systems} \bibinfo{volume}{29}, \bibinfo{pages}{3784--3797}.
\bibitem[{Douzas and Bacao(2019)}]{douzas2019geometric}
\bibinfo{author}{Douzas, G.}, \bibinfo{author}{Bacao, F.}, \bibinfo{year}{2019}.
\newblock \bibinfo{title}{Geometric smote a geometrically enhanced drop-in replacement for smote}.
\newblock \bibinfo{journal}{Information sciences} \bibinfo{volume}{501}, \bibinfo{pages}{118--135}.
\bibitem[{Douzas et~al.(2018)Douzas, Bacao and Last}]{douzas2018improving}
\bibinfo{author}{Douzas, G.}, \bibinfo{author}{Bacao, F.}, \bibinfo{author}{Last, F.}, \bibinfo{year}{2018}.
\newblock \bibinfo{title}{Improving imbalanced learning through a heuristic oversampling method based on k-means and smote}.
\newblock \bibinfo{journal}{Information sciences} \bibinfo{volume}{465}, \bibinfo{pages}{1--20}.
\bibitem[{Elreedy et~al.(2024)Elreedy, Atiya and Kamalov}]{elreedy2024distribution}
\bibinfo{author}{Elreedy, D.}, \bibinfo{author}{Atiya, A.F.}, \bibinfo{author}{Kamalov, F.}, \bibinfo{year}{2024}.
\newblock \bibinfo{title}{A theoretical distribution analysis of synthetic minority oversampling technique (smote) for imbalanced learning}.
\newblock \bibinfo{journal}{Machine Learning} \bibinfo{volume}{113}, \bibinfo{pages}{4903--4923}.
\bibitem[{Fern{\'a}ndez et~al.(2018a)Fern{\'a}ndez, Garc{\'\i}a, Galar, Prati, Krawczyk and Herrera}]{fernandez2018learning}
\bibinfo{author}{Fern{\'a}ndez, A.}, \bibinfo{author}{Garc{\'\i}a, S.}, \bibinfo{author}{Galar, M.}, \bibinfo{author}{Prati, R.C.}, \bibinfo{author}{Krawczyk, B.}, \bibinfo{author}{Herrera, F.}, \bibinfo{year}{2018}a.
\newblock \bibinfo{title}{Learning from imbalanced data sets}. volume~\bibinfo{volume}{10}.
\newblock \bibinfo{publisher}{Springer}.
\bibitem[{Fern{\'a}ndez et~al.(2018b)Fern{\'a}ndez, Garcia, Herrera and Chawla}]{fernandez2018smote}
\bibinfo{author}{Fern{\'a}ndez, A.}, \bibinfo{author}{Garcia, S.}, \bibinfo{author}{Herrera, F.}, \bibinfo{author}{Chawla, N.V.}, \bibinfo{year}{2018}b.
\newblock \bibinfo{title}{Smote for learning from imbalanced data: progress and challenges, marking the 15-year anniversary}.
\newblock \bibinfo{journal}{Journal of artificial intelligence research} \bibinfo{volume}{61}, \bibinfo{pages}{863--905}.
\bibitem[{Han et~al.(2005)Han, Wang and Mao}]{han2005borderline}
\bibinfo{author}{Han, H.}, \bibinfo{author}{Wang, W.Y.}, \bibinfo{author}{Mao, B.H.}, \bibinfo{year}{2005}.
\newblock \bibinfo{title}{Borderline-smote: a new over-sampling method in imbalanced data sets learning}, in: \bibinfo{booktitle}{International conference on intelligent computing}, \bibinfo{organization}{Springer}. pp. \bibinfo{pages}{878--887}.
\bibitem[{He et~al.(2008)He, Bai, Garcia and Li}]{he2008adasyn}
\bibinfo{author}{He, H.}, \bibinfo{author}{Bai, Y.}, \bibinfo{author}{Garcia, E.A.}, \bibinfo{author}{Li, S.}, \bibinfo{year}{2008}.
\newblock \bibinfo{title}{Adasyn: Adaptive synthetic sampling approach for imbalanced learning}, in: \bibinfo{booktitle}{2008 IEEE international joint conference on neural networks (IEEE world congress on computational intelligence)}, \bibinfo{organization}{Ieee}. pp. \bibinfo{pages}{1322--1328}.
\bibitem[{He and Garcia(2009)}]{he2009learning}
\bibinfo{author}{He, H.}, \bibinfo{author}{Garcia, E.A.}, \bibinfo{year}{2009}.
\newblock \bibinfo{title}{Learning from imbalanced data}.
\newblock \bibinfo{journal}{IEEE Transactions on knowledge and data engineering} \bibinfo{volume}{21}, \bibinfo{pages}{1263--1284}.
\bibitem[{Hemmatian et~al.(2025)Hemmatian, Hajizadeh and Nazari}]{hemmatian2025crn}
\bibinfo{author}{Hemmatian, J.}, \bibinfo{author}{Hajizadeh, R.}, \bibinfo{author}{Nazari, F.}, \bibinfo{year}{2025}.
\newblock \bibinfo{title}{Addressing imbalanced data classification with cluster-based reduced noise smote}.
\newblock \bibinfo{journal}{PLOS ONE} \bibinfo{volume}{20}, \bibinfo{pages}{e0317396}.
\bibitem[{Khalilia et~al.(2011)Khalilia, Chakraborty and Popescu}]{khalilia2011predicting}
\bibinfo{author}{Khalilia, M.}, \bibinfo{author}{Chakraborty, S.}, \bibinfo{author}{Popescu, M.}, \bibinfo{year}{2011}.
\newblock \bibinfo{title}{Predicting disease risks from highly imbalanced data using random forest}.
\newblock \bibinfo{journal}{BMC medical informatics and decision making} \bibinfo{volume}{11}, \bibinfo{pages}{51}.
\bibitem[{Khorshidi and Aickelin(2025)}]{khorshidi2025somm}
\bibinfo{author}{Khorshidi, H.A.}, \bibinfo{author}{Aickelin, U.}, \bibinfo{year}{2025}.
\newblock \bibinfo{title}{A synthetic over-sampling method with minority and majority classes for imbalance problems}.
\newblock \bibinfo{journal}{Knowledge and Information Systems} \bibinfo{volume}{67}, \bibinfo{pages}{5965--5998}.
\bibitem[{Krawczyk(2016)}]{krawczyk2016learning}
\bibinfo{author}{Krawczyk, B.}, \bibinfo{year}{2016}.
\newblock \bibinfo{title}{Learning from imbalanced data: open challenges and future directions}.
\newblock \bibinfo{journal}{Progress in artificial intelligence} \bibinfo{volume}{5}, \bibinfo{pages}{221--232}.
\bibitem[{Li et~al.(2025)Li, Yang, Song, Duan and Ren}]{li2025ismote}
\bibinfo{author}{Li, Y.}, \bibinfo{author}{Yang, Y.}, \bibinfo{author}{Song, P.}, \bibinfo{author}{Duan, L.}, \bibinfo{author}{Ren, R.}, \bibinfo{year}{2025}.
\newblock \bibinfo{title}{An improved smote algorithm for enhanced imbalanced data classification by expanding sample generation space}.
\newblock \bibinfo{journal}{Scientific Reports} \bibinfo{volume}{15}, \bibinfo{pages}{23521}.
\bibitem[{Matharaarachchi et~al.(2024)Matharaarachchi, Domaratzki and Muthukumarana}]{matharaarachchi2024extsmote}
\bibinfo{author}{Matharaarachchi, S.}, \bibinfo{author}{Domaratzki, M.}, \bibinfo{author}{Muthukumarana, S.}, \bibinfo{year}{2024}.
\newblock \bibinfo{title}{Enhancing smote for imbalanced data with abnormal minority instances}.
\newblock \bibinfo{journal}{Machine Learning with Applications} \bibinfo{volume}{18}, \bibinfo{pages}{100597}.
\bibitem[{Nikpour et~al.(2026)Nikpour, Rahmati, Mirzaei and Nezamabadi-pour}]{nikpour2026review}
\bibinfo{author}{Nikpour, B.}, \bibinfo{author}{Rahmati, F.}, \bibinfo{author}{Mirzaei, B.}, \bibinfo{author}{Nezamabadi-pour, H.}, \bibinfo{year}{2026}.
\newblock \bibinfo{title}{A comprehensive review on data-level methods for imbalanced data classification}.
\newblock \bibinfo{journal}{Expert Systems with Applications} \bibinfo{volume}{295}, \bibinfo{pages}{128920}.
\bibitem[{Qiu et~al.(2025)Qiu, Yang, Xiong, Chen, Xiao, Zheng, Chen and Zhou}]{qiu2025vs}
\bibinfo{author}{Qiu, F.}, \bibinfo{author}{Yang, F.}, \bibinfo{author}{Xiong, Y.}, \bibinfo{author}{Chen, T.}, \bibinfo{author}{Xiao, P.}, \bibinfo{author}{Zheng, W.}, \bibinfo{author}{Chen, X.}, \bibinfo{author}{Zhou, K.}, \bibinfo{year}{2025}.
\newblock \bibinfo{title}{Vs-smote: Leveraging high-value spaces to balance noise control and data diversity for class imbalanced learning}.
\newblock \bibinfo{journal}{Neurocomputing} , \bibinfo{pages}{131321}.
\bibitem[{Sowjanya and Mrudula(2023)}]{sowjanya2023effective}
\bibinfo{author}{Sowjanya, A.M.}, \bibinfo{author}{Mrudula, O.}, \bibinfo{year}{2023}.
\newblock \bibinfo{title}{Effective treatment of imbalanced datasets in health care using modified smote coupled with stacked deep learning algorithms}.
\newblock \bibinfo{journal}{Applied Nanoscience} \bibinfo{volume}{13}, \bibinfo{pages}{1829--1840}.
\bibitem[{Sun et~al.(2025)Sun, Li and Zhu}]{sun2025ebsmote}
\bibinfo{author}{Sun, H.}, \bibinfo{author}{Li, J.}, \bibinfo{author}{Zhu, X.}, \bibinfo{year}{2025}.
\newblock \bibinfo{title}{A novel expandable borderline smote oversampling method for class imbalance problem}.
\newblock \bibinfo{journal}{IEEE Transactions on Knowledge and Data Engineering} .
\bibitem[{Sun et~al.(2024)Sun, Wang, Jia and Xu}]{sun2024smote}
\bibinfo{author}{Sun, P.}, \bibinfo{author}{Wang, Z.}, \bibinfo{author}{Jia, L.}, \bibinfo{author}{Xu, Z.}, \bibinfo{year}{2024}.
\newblock \bibinfo{title}{Smote-ktlnn: A hybrid re-sampling method based on smote and a two-layer nearest neighbor classifier}.
\newblock \bibinfo{journal}{Expert Systems with Applications} \bibinfo{volume}{238}, \bibinfo{pages}{121848}.
\bibitem[{Taskiran et~al.(2025)Taskiran, Turkoglu, Kaya and Asuroglu}]{taskiran2025evaluation}
\bibinfo{author}{Taskiran, S.F.}, \bibinfo{author}{Turkoglu, B.}, \bibinfo{author}{Kaya, E.}, \bibinfo{author}{Asuroglu, T.}, \bibinfo{year}{2025}.
\newblock \bibinfo{title}{A comprehensive evaluation of oversampling techniques for enhancing text classification performance}.
\newblock \bibinfo{journal}{Scientific Reports} \bibinfo{volume}{15}, \bibinfo{pages}{21631}.
\bibitem[{Udu et~al.(2025)Udu, Salman, Ghalati, Lecchini-Visintini, Siddle and Dong}]{udu2025emerging}
\bibinfo{author}{Udu, A.G.}, \bibinfo{author}{Salman, M.T.}, \bibinfo{author}{Ghalati, M.K.}, \bibinfo{author}{Lecchini-Visintini, A.}, \bibinfo{author}{Siddle, D.R.}, \bibinfo{author}{Dong, H.}, \bibinfo{year}{2025}.
\newblock \bibinfo{title}{Emerging smote and gan-variants for data augmentation in imbalance machine learning tasks: A review}.
\newblock \bibinfo{journal}{IEEE Access} .
\bibitem[{Wang et~al.(2025)Wang, Zheng, Ma and Hu}]{wang2025resampling}
\bibinfo{author}{Wang, F.}, \bibinfo{author}{Zheng, M.}, \bibinfo{author}{Ma, K.}, \bibinfo{author}{Hu, X.}, \bibinfo{year}{2025}.
\newblock \bibinfo{title}{Resampling approach for imbalanced data classification based on class instance density per feature value intervals}.
\newblock \bibinfo{journal}{Information Sciences} \bibinfo{volume}{692}, \bibinfo{pages}{121570}.
\bibitem[{Wang et~al.(2026)Wang, Mohd~Rosli and Musa}]{wang2026overlap}
\bibinfo{author}{Wang, Y.}, \bibinfo{author}{Mohd~Rosli, M.}, \bibinfo{author}{Musa, N.}, \bibinfo{year}{2026}.
\newblock \bibinfo{title}{Class overlap in imbalanced learning: A data-level perspective and comprehensive review}.
\newblock \bibinfo{journal}{Journal of King Saud University Computer and Information Sciences} .
\bibitem[{Widiyaningtyas et~al.(2025)Widiyaningtyas, Hairani, Prasetya, Pujianto and Caesarendra}]{widiyaningtyas2025nrmodified}
\bibinfo{author}{Widiyaningtyas, T.}, \bibinfo{author}{Hairani, H.}, \bibinfo{author}{Prasetya, D.D.}, \bibinfo{author}{Pujianto, U.}, \bibinfo{author}{Caesarendra, W.}, \bibinfo{year}{2025}.
\newblock \bibinfo{title}{A modified smote with noise filtering and manhattan distance metric approach to address imbalanced health datasets}.
\newblock \bibinfo{journal}{Engineering, Technology \& Applied Science Research} \bibinfo{volume}{15}, \bibinfo{pages}{25452--25459}.
\bibitem[{Zhou et~al.(2022)Zhou, Hu, Wu, Liang, Ma and Jin}]{zhou2022distribution}
\bibinfo{author}{Zhou, X.}, \bibinfo{author}{Hu, Y.}, \bibinfo{author}{Wu, J.}, \bibinfo{author}{Liang, W.}, \bibinfo{author}{Ma, J.}, \bibinfo{author}{Jin, Q.}, \bibinfo{year}{2022}.
\newblock \bibinfo{title}{Distribution bias aware collaborative generative adversarial network for imbalanced deep learning in industrial iot}.
\newblock \bibinfo{journal}{IEEE Transactions on Industrial Informatics} \bibinfo{volume}{19}, \bibinfo{pages}{570--580}.

\end{thebibliography}

\end{document}